\documentclass[a4paper,conference]{IEEEtran}
\usepackage{times}
\usepackage{epsfig}
\usepackage{graphicx}
\usepackage{amsmath}
\usepackage{amssymb}
\usepackage{pifont}
\usepackage{multirow}

\usepackage[bookmarks=false,colorlinks]{hyperref}
\usepackage[frozencache=true,cachedir=mintedcache]{minted}

\begin{document}

\title{Unified Recurrence Modeling for Video Action Anticipation}

\makeatletter
\newcommand{\linebreakand}{%
  \end{@IEEEauthorhalign}
  \hfill\mbox{}\par
  \mbox{}\hfill\begin{@IEEEauthorhalign}
}
\makeatother
\author{
\IEEEauthorblockN{Tsung-Ming Tai}
\IEEEauthorblockA{NVIDIA AI Technology Center}
\IEEEauthorblockA{Free University of Bozen-Bolzano\\
ntai@nvidia.com, tstai@unibz.it}
\and
\IEEEauthorblockN{Giuseppe Fiameni}
\IEEEauthorblockA{NVIDIA AI Technology Center\\
gfiameni@nvidia.com}
\and
\IEEEauthorblockN{Cheng-Kuang Lee}
\IEEEauthorblockA{NVIDIA AI Technology Center\\
cklee@nvidia.com}
\linebreakand
\IEEEauthorblockN{Simon See}
\IEEEauthorblockA{NVIDIA AI Technology Center\\
ssee@nvidia.com}
\and
\IEEEauthorblockN{Oswald Lanz}
\IEEEauthorblockA{Free University of Bozen-Bolzano\\
lanz@inf.unibz.it}
}

\newcommand{\revgf}[2]{\textcolor{purple}{#2}}
\newcommand{\revnick}[2]{\textcolor{brown}{#2}}

\maketitle

\begin{abstract}
Forecasting future events based on evidence of current conditions is an innate skill of human beings, and key for predicting the outcome of any decision making. In artificial vision for example, we would like to predict the next human action before it happens, without observing the future video frames associated to it. Computer vision models for action anticipation are expected to collect the subtle evidence in the preamble of the target actions. In prior studies recurrence modeling often leads to better performance, the strong temporal inference is assumed to be a key element for reasonable prediction. To this end, we propose a unified recurrence modeling for video action anticipation via message passing framework. The information flow in space-time can be described by the interaction between vertices and edges, and the changes of vertices for each incoming frame reflects the underlying dynamics. Our model leverages self-attention as the building blocks for each of the message passing functions. In addition, we introduce different edge learning strategies that can be end-to-end optimized to gain better flexibility for the connectivity between vertices. Our experimental results demonstrate that our proposed method outperforms previous works on the large-scale EPIC-Kitchen dataset.
\end{abstract}

\section{Introduction}
\begin{figure*}[!t]
    \centering
    \includegraphics[width=1\textwidth]{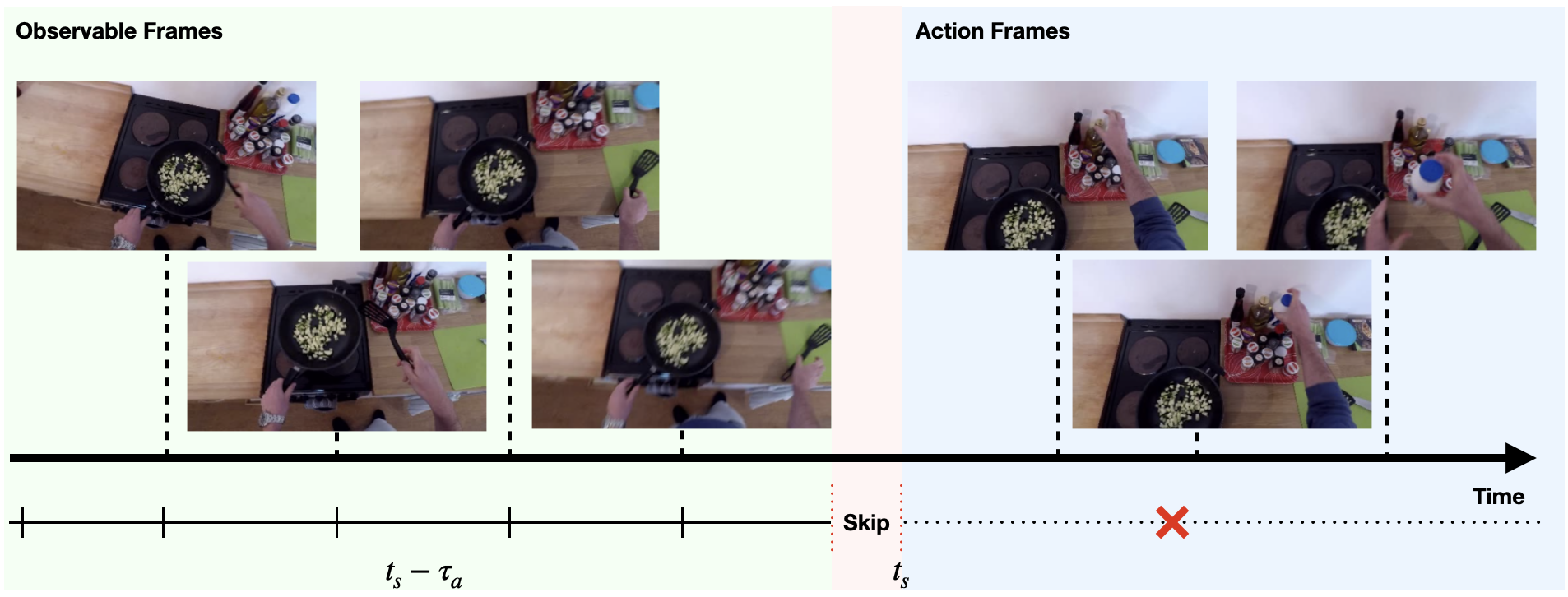}
    \caption{Illustration of video action anticipation problem. The context of action may be different than the preamble of action. Models can only observe some frames before action actually starts $t_s$, which is strictly ensured by an inaccessible "skip" period, and based on the evidence collected in the duration in $t_s - \tau_a$ to predict the next action. The anticipation accuracy is measured at each different $\tau_a$.}
    \label{fig:video_anticipation}
\end{figure*}

Video action recognition is a long-standing problem in computer vision. The goal is to predict the action category that can be observed in a video clip, by recognizing the action discriminant spatio-temporal patterns and their context in the observations~\cite{DBLP:journals/corr/abs-2012-06567}. When it comes to action anticipation, however, a prediction must be made {\em before} the actual action is observed \cite{furnari2018leveraging, furnari2020rulstm, rodin2021predicting}. This is relevant for many real-world applications, for example, in assistive navigation systems \cite{ohnbar2018personalized}, collaborative robotics \cite{park2016egocentric}, interactive entertainment \cite{liang2015ar, taylor2020towards} and autonomous vehicles \cite{hirakawa2018survey}. Figure~\ref{fig:video_anticipation} illustrates the definition of the video anticipation problem. Naively re-framing clip-based action recognition to perform anticipation is inadequate from a model design perspective, and may result in inefficiency.

Recurrent neural networks are widely adopted for modeling the temporal relationship in anticipation problems \cite{furnari2020rulstm,9356220,osman2021slowfast} and lead to better performance than clip-based methods \cite{atsn,drm}, as opposed to the mainstream in action recognition. The basic assumption is that in action anticipation the observations are incomplete and indirect, although related, to the target action. Some actions also come with misleading and unclear preamble information, for example, there are many future possibilities when observing someone extending its hand or moving closer to an area with many actionable objects in sight. Action anticipation is more towards forecasting than recognition, and thus the effectiveness of adapting action recognition models to the anticipation problem is reduced. 

In this paper we present a unified recurrence modeling for video action anticipation which generalizes the recurrence mechanism by transferring the sequence learning into a graph representation learning realized via a message passing framework. We use self-attention as the universal building block for extracting information in vertices and edges. Vertices are associated to the representations provided by a backbone's specific receptive field. Edges describe the bonding strength between the vertices. Self-attention can be seen as the function of routing information between vertices. The attention weights are derived by the scaled dot-product, which computes the correlation of vertices, and can be interpreted as an implicit adjacency estimation. However, in this way, the representation of edges is limited and purely based on the similarity between vertices. To improve such trivial estimation, we propose edge learning strategies which explicitly approximate the edge connectivity. In the experimental results, we show that the proposed unified recurrent modeling outperforms several state-of-the-art methods on the large-scale egocentric video dataset EPIC-Kitchens. When combined with edge learning strategies, we obtain a further significant boost in performance.

\section{Related work}
\subsection{Video Action Anticipation}
Early works in video anticipation model the problem with recurrent neural networks \cite{DBLP:conf/bmvc/GaoYN17a, DBLP:conf/cvpr/FarhaRG18, DBLP:conf/cvpr/MiechLSWTT19}. Some prior works also leverage the future frames for learning the representations \cite{DBLP:conf/cvpr/VondrickPT16, fernando2021anticipating}. A self-regulated learning framework for action anticipation on the egocentric video is presented in~\cite{9356220}, which learns to emphasize the important context by its revision and reattend designs. RU-LSTM \cite{DBLP:conf/iccv/FurnariF19} deploys two LSTMs and behaves as an encoder-decoder, where the first progressively summarizes the observed together with the second that unrolls over future predictions without observing. The unrolling design can also be found in \cite{osman2021slowfast, tai2021higher}, but with the rolling part replaced by SlowFast \cite{feichtenhofer2019slowfast} and Higher-Order Recurrent Transformer, respectively. \cite{sener2020temporal} aggregates the multiple predictions by pooling over different granularity of temporal segments to improve the anticipation accuracy. \cite{girdhar2021anticipative} combines causal self-attention with several regularization terms, showing the strong performance on the video anticipation problems. Our method also utilizes self-attention and unrolling designs. In addition, the proposed model learns how to propagate information in space-time via the generic message passing framework. All latent representations are further processed and contextualized with information from vertices and edges.
 
\subsection{Message Passing Neural Network}
The concept of message passing in neural networks is introduced in \cite{gilmer2017neural}, where it was originally designed for molecular property prediction. It assumes an undirected graph structure with data-independent, equal edge contributions. To address this limitation, \cite{ma2020dual} deploys two encoders separately for vertex and edge estimation and aggregates them by an attention readout. Similar works also leverage attention or dedicated network design to learn the directed edge representations to improve the model capability \cite{jorgensen2018neural, withnall2020building, gong2019exploiting}. Recently, \cite{arnab2021unified} reinterpreted the Non-Local \cite{DBLP:conf/cvpr/0004GGH18} and GAT \cite{velivckovic2017graph} as a message passing functions and apply them to video understanding task. Differently, we view the message passing framework as the generalized recurrent models and specialize it for edge representations learning.

\subsection{Self-Attention}
\cite{DBLP:conf/nips/VaswaniSPUJGKP17} first proposed a recurrence-free sequence learning architecture by stacking several self-attention layers, which can achieve remarkable performance in the NLP domain. \cite{dehghani2018universal} demonstrates that self-attention can be treated as the recurrent unit which unfolds to input sequences to processes with shared weights.
On the other hand, \cite{dosovitskiy2020image} proposes Vision Transformer (ViT), an architecture with only self-attention for image classification. ViT inherits the class-token design from \cite{devlin2018bert} to better represent the prediction hypothesis. \cite{zhou2021deepvit} further improves ViT by re-attending the multi-heads information in the post-softmax step to enable deeper configuration. Some recent studies explore ViT based models on video action recognition \cite{DBLP:journals/corr/abs-2103-15691, DBLP:journals/corr/abs-2102-05095}, and also video anticipation \cite{girdhar2021anticipative}. Unlike these prior works, our proposed model processes the video in a flexible graph representation and leverages the message passing framework. Our model is lightweight and only contains a few self-attention layers which sequentially process each timestep.

\section{Method}
\subsection{Backgrounds}
\subsubsection{Message Passing}
Given an undirected graph $G$, the Message Passing algorithm involves a two-phase forwarding process. It is composed of (i) message passing phase with message function $M$ and update function $U$; and (ii) readout phase with readout function $R$. The message passing phase can be executed in an arbitrary number of $T$ steps, where each time step learns to update the graph representation according to the following definition,
\begin{eqnarray}
    m_v^{t+1} &=& \sum_{w \in N(v)} M_t (h_v^t, h_w^t, a_{vw}) \label{eq:mpnn_msgfn}\\
    h_v^{t+1} &=& U_t(h_v^t, m_v^{t+1}) \label{eq:mpnn_updatefn}
\end{eqnarray}
where $v$ is the vertices in $G$, and $N(v)$ defines the neighbors of $v$. $a_{vw}$ is the connection strength bonding between vertex $v$ and $w$. The readout phase then extracts the features of the whole graph produced in the message passing phase, at the final time step $T$,
\begin{eqnarray}
    \hat{y} &=& R({h_v^T | v \in G}) \label{eq:mpnn_readoutfn}.
\end{eqnarray}
Our method inherits these three core functions, i.e., message $M$, update $U$, and readout $R$. 
 
\subsubsection{Self-Attention}
Self-Attention (SA) forms the $Q,K,V$ tokens (for query, key, and value) from the input $x$. The output of attention is a weighted linear combination of values. Attention weights are computed by the scaled dot-product between query and key followed by a softmax,
\begin{eqnarray}
    Q, K, V &=& xW_Q^i, xW_K^i, xW_V^i \\
    SA_i(x) &=& softmax(\frac{Q^T K}{\sqrt{D}})V, \label{eq:sa}
\end{eqnarray}
where scaling factor $D$ is the input feature dimension, and $W_Q^i,W_K^i,W_V^i$ are the trainable embeddings. We use superscript $i$ to indicate that the embeddings are associated with each self-attention layer.

Multi-Head Self Attention (MHSA) performs $n$-way self-attention in parallel, where $n$ is the total number of heads. An additional aggregation function, with parameters $W_{agg}$, is adopted to fuse the information computed from each head,
\begin{eqnarray}
    MHSA(x) &=& [H_1,\dots,H_n] W_{agg}, \label{eq:mhsa}\\
    where~H_i &=& SA_i(x) \nonumber
\end{eqnarray}
where $[.,.]$ presents concatenation.
 
Following the transformer-style architecture, a Feed-Forward Network (FFN) is introduced after each attention layer to project the attention output and bring the non-linearity. The FFN computes
\begin{eqnarray}
    FFN(x) = \sigma(x W_1 + b_1) W_2 + b_2 \label{eq:ffn}
\end{eqnarray}
where $\sigma$ can be any arbitrary nonlinear function.

\subsection{Unified Recurrent Modeling}
\subsubsection{Self-Attention Block (SABlock)}
Following \cite{xiong2020layer}, as with many transformer-style architectures, we define our self-attention building block (SABlock) in prenorm style. It contains a Multi-Head Self-Attention (MHSA) followed by a FFN.
Figure~\ref{fig:overview} left shows the SABlock design,
\begin{eqnarray}
    f_{MHSA}(x) &=& x + MHSA(LayerNorm(x)) \\
    f_{FFN}(x) &=& x + FFN(LayerNorm(x)) \\
    SABlock(x) &=& f_{FFN}(f_{MHSA}(x)) \label{eq:SABlock}
\end{eqnarray}
We can optionally expose the edge information, in the form of the adjacency matrix $A$, into $SABlock(.; A)$ by fusing $A$ into the step after the softmax of scaled dot-product computation. The extension can also be applied to $MHSA(x; A)$. Accordingly, we rewrite~\eqref{eq:sa} and~\eqref{eq:mhsa} as
\begin{eqnarray}
    MHSA(x; A) &=& [H_1,\dots,H_n] W_{agg}, \label{eq:extend_mhsa}\\
    where~H_i &=& SA_i(x; A) \nonumber \\
    SA_i(x; A) &=& \Bigg( softmax(A) + softmax(\frac{Q^TK}{\sqrt{D}}) \Bigg)V \nonumber\\ 
    \label{eq:extend_sa}
\end{eqnarray}
Note that $A$ is unique and shared in multi-heads self-attention.

\begin{figure}
    \centering
    \includegraphics[width=0.48\textwidth]{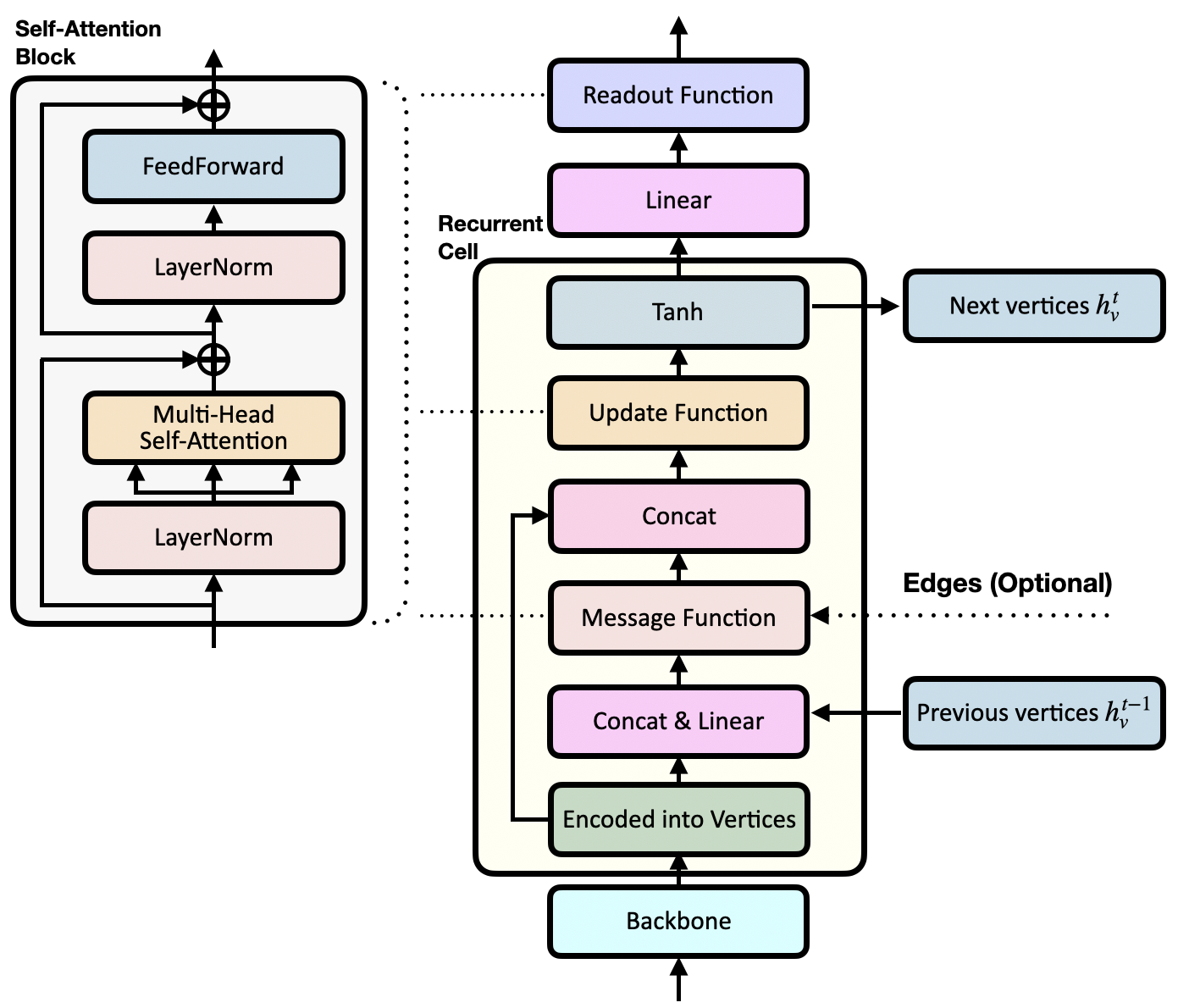}
    \caption{Overview of proposed unified recurrent model. The message function, update function, and readout function leverage multi-head self-attention. Our proposed modeling scheme is also flexible to work in conjugation with explicit edges information provided.}
    \label{fig:overview}
\end{figure}

\begin{figure*}
    \centering
    \includegraphics[width=0.91\textwidth]{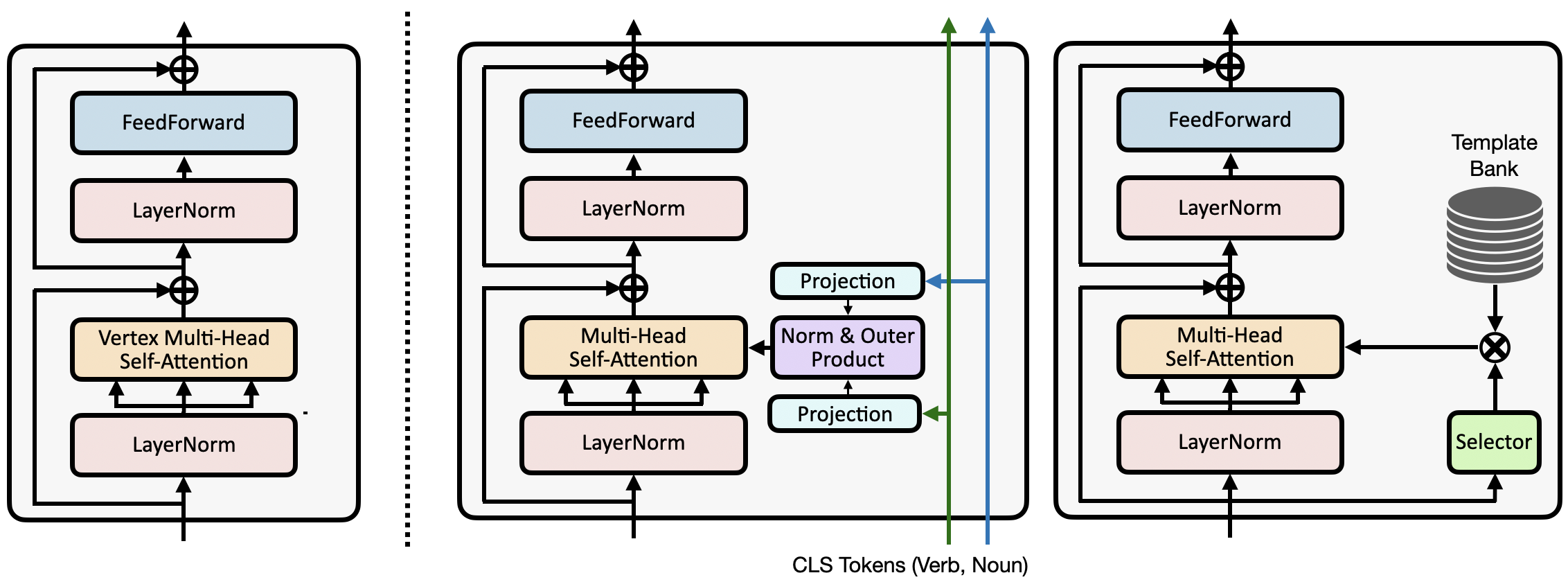}
    \caption{Proposed edge learning extensions to the multi-head self-attention layer. The \textit{left} figure shows the original self-attention block; The two figures at the \textit{right} illustrate two proposed edge learning strategies, Class-Token Projection (CTP) and Template Bank (TB), respectively.}
    \label{fig:edge_learning}
\end{figure*}

\subsubsection{Recurrent Cell}
Figure~\ref{fig:overview} shows the design of the proposed architecture. Given the frame features $x^t$ at time $t$ in shape ($H$, $W$, $C$), computed by a frame-level feature extractor, we compute the vertices $e_v^t$ by a gated nonlinear transformation of $x^t$. The vertex representations are reshaped into ($HW$, $C$), which presents the number of vertices $HW$ and feature dimension $C$ of each vertex. A position encoding $W_{pe}$ is added to $e_v^t$. We then leverage $SABlock(.; A)$ with the adjacency matrix $A$ for modeling each core function: message function, update function, and readout function.

The message function is a Multi-Layer Perceptron (MLP) with a self-gated design to form the vertex representations $e_v^t$. We concatenate the information coming from the previous $h_v^{t-1}$ with an additional linear transformation. The output is then processed by the self-attention block. Overall,
\begin{eqnarray}
    \overline{x^t} &=& MLP(x^t) \\
    e_v^{t} &=& sigmoid( \overline{x^t}) \cdot  \overline{x^t} + W_{pe} \label{eq:our_enc fn} \\
    g_v^{t} &=& W_h([e_v^t, h_v^{t-1}]) + b_h \label{eq:our_encfn2} \\
    m_v^{t} &=& SABlock(g_v^{t}; A^t) \label{eq:our_msgfn}
\end{eqnarray}  
Note that the linear transform, parameterized by $W_h$ and $b_h$, reduces the feature size from $2C$ to $C$.

The update function is another self-attention block that updates the graph representation from the message function with the vertices derived from the current frame,
\begin{eqnarray}
    u_v^{t} = SABlock([e_v^t, m_v^{t}]). \label{eq:our_updatefn}
\end{eqnarray}
For better stability during the temporal propagation, the hyperbolic tangent (tanh) is applied to the computed graph representation from the update function,
\begin{eqnarray}
    h_v^{t} = tanh(u_v^{t}). \label{eq:our_updatefn2}
\end{eqnarray}

The readout function, which is also formed by self-attention, retrieves the layer output from the updated graph representation $h_v^{t}$,
\begin{eqnarray}
    y^{t} = SABlock(W_r h_v^{t} + b_r) \label{eq:our_readoutfn}
\end{eqnarray}
where $W_r$ and $b_r$ are the weights of a linear transformation that decodes the hidden representation, which is bounded to [-1, 1], into a more flexible value range. The output $y^{t}$ is then sent to a task-specific classifier.

Our design translates the anticipation problem into a message passing scheme producing a graph-structured space-time representation. The connectivity of the graph structure is inferred from the input at each time step, as described next.  The readout function is called when the prediction is required at any time $t$. The proposed model utilizes only multi-head self-attention for information routing between vertices. Note that the resulting spatial graph is either bi-directed, when an adjacency matrix $A$ is provided, or else it is undirected.

Without any prior knowledge, we assume each vertex in the graph $G$ is accessible by all other vertices ($N(v)$ in \eqref{eq:mpnn_msgfn} contains all vertices). In this case the scaled dot-product in the self-attention computes the pairwise similarity of all vertices from the inputs, which can be viewed as an {\em implicit} edge estimation. This can be extended by optionally providing the edge estimation {\em explicitly}, in the form of an adjacency matrix $A=\{a_{vw}; \forall w \in N(v), v \in G\}$, and 
using it
during the attention computation as described in~\eqref{eq:extend_mhsa} and \eqref{eq:extend_sa}. We refer to these two cases as \textit{implicit} and \textit{explicit} edge learning.

\subsection{Explicit Edge Learning} \label{sec:edge_learning}
We propose two learning strategies to explicitly construct the edge information, as shown in Figure~\ref{fig:edge_learning}. Template Bank (TB) forms the estimation of edge connections by soft-fusing a set of learnable templates using weights computed from the frame input. Class Token Projection (CTP) performs the outer-product of class tokens to construct the edge estimation, which are supervised from provided class labels. Both methods operate at frame-level at time $t$ and approximate the corresponding edge estimation $\widehat{A}^t$ for the message function in~\eqref{eq:our_msgfn}.

\subsubsection{Template bank (TB)}
We introduce a template bank $B$ of size $S$. Each template in the bank is of shape ($N$, $N$) where $N$ is the number of vertices in our graph. During the forward process, the adjacency matrix  $\widehat{A}^t$ is formed by a weighted sum over the $S$ templates. Template weights $I$ are computed from an aggregated representation of frame inputs using a selector $f_{select}$, consisting of a multi-layer perception (MLP), followed by softmax normalization,
\begin{eqnarray}
    \overline{e_v^t} &=& \frac{1}{N}\sum_{i=0}^N e_{v,i}^t \\
    I &=& softmax(f_{select}(\overline{e_v^t})) \\
    \widehat{A}^t &=& \sum_{i=0}^{S} I_i \cdot B_{i, :, :}
\end{eqnarray}
where $\overline{e_v^t}$ is the aggregated representation of the input obtained by average pooling of vertex information $e_v^t$.

\subsubsection{Class Token Projection (CTP)}
Leveraging the fact that the class tokens receive the supervision signal from ground-truth labeling, we can span the edge estimation by adopting the outer-product of class tokens projected into an embedding space. The learnable class tokens $\texttt{CLS}_{verb}^t$ and $\texttt{CLS}_{noun}^t$ represent the \textit{verb} and \textit{noun} hypothesis at time $t$, and have dimension ($1$, $C$). We project both class tokens individually by introducing two linear transforms that map each token into an embedding space whose dimension matches the number of vertices $N$. The outer-product of the projected token vectors then defines the adjacency matrix $\widehat{A}^t$ of shape ($N$, $N$),
\begin{eqnarray}
    V &=& LN(W_v \texttt{CLS}_{verb}^t + b_v) \\
    N &=& LN(W_n \texttt{CLS}_{noun}^t + b_n) \\
    \widehat{A}^t &=& V \otimes N,\label{eq:class-token projection}
\end{eqnarray}
where $LN$ is the layer normalization, $\otimes$ is the outer-product operator, and $W_v,W_n,b_v,b_n$ are the learnable weights of the projection functions. Note that with this definition, the class tokens are directly used in the self-attention computation of the readout function and in the task-specific classifier.

\section{Experimental Results}
\begin{table*}[!ht]
\caption{EK55 Action Anticipation validation results using RGB at different $\tau_a$ in Top-5 action accuracy.}
    \centering
    \scriptsize
    \begin{tabular}{l|cccccccc|ccc|ccc}
    \multirow{2}{*}{Methods} &
    \multicolumn{8}{c}{Top-5 Action Accuracy (\%) at different $\tau_a$} &
    \multicolumn{3}{c}{Top-5 Acc. (\%) @ 1\textit{s}} &
    \multicolumn{3}{c}{Mean Top-5 Rec. (\%) @ 1\textit{s}}\\
    & 2.00 & 1.75 & 1.50 & 1.25 & 1.00 & 0.75 & 0.50 & 0.25 &
    Verb & Noun & Action & Verb & Noun & Action\\
    \hline
    
    DMR \cite{DBLP:conf/cvpr/VondrickPT16} & - & - & - & - & 16.86 & - & - & - & 73.66 & 29.99 & 16.86 & 24.50 & 20.89 & 03.23 \\
    ATSN \cite{atsn} & - & - & - & - & 16.29 & - & - & - & 77.30 & 39.93 & 16.29 & 33.08 & 32.77 & 07.06 \\
    MCE \cite{furnari2018leveraging} & - & - & - & - & 26.11 & - & - & - & 73.35 & 38.86 & 26.11 & 34.62 & 32.59 & 06.50 \\
    VN-CE \cite{furnari2018leveraging} & - & - & - & - & 17.31 & - & - & - & 77.67 & 39.50 & 17.31 & 34.05 & 34.50 & 07.73 \\
    SVM-TOP3 \cite{berrada2018smooth} & - & - & - & - & 25.42 & - & - & - & 72.70 & 28.41 & 25.42 & 41.90 & 34.69 & 05.32 \\
    SVM-TOP5 \cite{berrada2018smooth} & - & - & - & - & 24.46 & - & - & - & 69.17 & 36.66 & 24.46 & 40.27 & 32.69 & 05.23 \\
    VNMCE+T3 \cite{furnari2018leveraging} & - & - & - & - & 25.95 & - & - & - & 74.05 & 39.18 & 25.95 & 40.17 & 34.15 & 05.57 \\
    VNMCE+T5 \cite{furnari2018leveraging} & - & - & - & - & 26.01 & - & - & - & 74.07 & 39.10 & 26.01 & 41.62 & 35.49 & 05.78 \\
    ED \cite{DBLP:conf/bmvc/GaoYN17a} & 21.53 & 22.22 & 23.20 & 24.78 & 25.75 & 26.69 & 27.66 & 29.74 & 75.46 & 42.96 & 25.75 & 41.77 & 42.59 & 10.97 \\
    FN \cite{DBLP:conf/wacv/GeestT18} & 23.47 & 24.07 & 24.68 & 25.66 & 26.27 & 26.87 & 27.88 & 28.96 & 74.84 & 40.87 & 26.27 & 35.30 & 37.77 & 06.64 \\
    RL \cite{DBLP:conf/cvpr/MaSS16} & 25.95 & 26.49 & 27.15 & 28.48 & 29.61 & 30.81 & 31.86 & 32.84 & 76.79 & 44.53 & 29.61 & 40.80 & 40.87 & 10.64 \\
    EL \cite{DBLP:conf/icra/JainSKSS16} & 24.68 & 25.68 & 26.41 & 27.35 & 28.56 & 30.27 & 31.50 & 33.55 & 75.66 & 43.72 & 28.56 & 38.70 & 40.32 & 08.62 \\
    RU-RGB \cite{furnari2020rulstm} & 25.44 & 26.89 & 28.32 & 29.42 & 30.83 & 32.00 & 33.31 & 34.47 & - & - & 30.83 & - & - & - \\
    SRL \cite{9356220} & 25.82 & 27.21 & 28.52 & 29.81 & 31.68 & 33.11 & 34.75 & 36.89 & \textbf{78.90} & 47.65 & 31.68 & 42.83 & 47.64 & 13.24 \\
    SF-RU \cite{osman2021slowfast} ($\alpha_s$=$\frac{1}{8}$) & 24.53 & 25.63 & 27.30 & 28.97 & 30.96 & 32.23 & 33.49 & 35.02 & - & - & 30.96 & - & - & - \\
    SF-RU \cite{osman2021slowfast} ($\alpha_s$=$\frac{1}{2}$) & 26.39 & - & 28.40 & - & 30.94 & - & 32.87 & - & - & - & 30.94 & - & - & - \\
    SF-RU \cite{osman2021slowfast} ($\alpha_s$=$\frac{1}{2},\frac{1}{8}$) & 26.78 & - & 29.25 & - & \textbf{32.05} & - & 34.34 & - & - & - & \textbf{32.05} & - & - & - \\
    \hline
    \textbf{Ours (Implicit)} & \textbf{27.25} & 27.76 & 29.36 & \textbf{30.63} & 31.68 & 32.76 & 34.41 & 36.65 & 78.66 & \textbf{47.93} & 31.68 & 43.67 & \textbf{47.93} & 13.19 \\
    \textbf{Ours (Explicit: TB)} & 26.67 & 27.76 & 29.32 & 30.49 & \textbf{32.02} & \textbf{33.47} & 34.71 & 36.85 & 78.60 & 46.86 & \textbf{32.02} & 43.63 & 46.86 & \textbf{13.58} \\
    \textbf{Ours (Explicit: CTP)} & 26.87 & \textbf{27.90} & \textbf{29.44} & \textbf{30.63} & 31.96 & 33.19 & \textbf{34.92} & \textbf{37.05} & 78.74 & 47.59 & 31.96 & \textbf{44.96} & 47.19 & \textbf{13.61} \\
    \end{tabular}
    \label{tab:ek55}
\end{table*}

\begin{table}[!ht]
\caption{EK55 Action Anticipation validation results using RGB with top-1 and top-5 action accuracy at $\tau_a = 1\textit{s}$.}
    \centering
    \scriptsize
    \begin{tabular}{l|l|l|cc}
         Method & Backbone & Pretrain & Top-1 (\%) & Top-5 (\%) \\
         \hline
         RU-RGB \cite{furnari2020rulstm} & \underline{BNInc} & \underline{In1k} & 13.1 & 30.8 \\
         ActionBanks \cite{sener2020temporal} & \underline{BNInc} & \underline{In1k} & 12.3 & 28.5 \\
         ImagineRNN \cite{wu2020learning} & \underline{BNInc} & \underline{In1k} & 13.7 & 31.6 \\
         AVT-h \cite{girdhar2021anticipative} & \underline{BNInc} & \underline{In1k} & 13.1 & 28.1 \\
         AVT-h \cite{girdhar2021anticipative} & AVT-b & In21k+1k & 12.5 & 30.1 \\
         AVT-h \cite{girdhar2021anticipative} & irCSN152 & IG65M & \textbf{14.4} & 31.7 \\
         \hline
         \textbf{Ours (Implicit)} & \underline{BNInc} & \underline{In1k} & 13.5 & 31.7 \\
         \textbf{Ours (Explicit: TP)} & \underline{BNInc} & \underline{In1k} & 13.8 & \textbf{32.0} \\
         \textbf{Ours (Explicit: CTP)} & \underline{BNInc} & \underline{In1k} & 13.6 & \textbf{32.0} \\
    \end{tabular}
    \label{tab:ek55_3}
\end{table}

\begin{table}[!ht]
\caption{EK100 Action Anticipation validation results using RGB with mean top-5 recall (\%) at $\tau_a = 1\textit{s}$.}
    \centering
    \scriptsize
    \begin{tabular}{l|l|l|ccc}
         Method & Backbone & Pretrain & Verb & Noun & Action \\
         \hline
         RU-RGB \cite{furnari2020rulstm} & \underline{BNInc} & \underline{In1k} & 27.5 & 29.0 & 13.3 \\
         AVT-h \cite{girdhar2021anticipative} & \underline{BNInc} & \underline{In1k} & 27.3 & 30.7 & 13.6 \\
         AVT-h \cite{girdhar2021anticipative} & AVT-b & In1k & 28.2 & 29.3 & 13.4 \\
         AVT-h \cite{girdhar2021anticipative} & AVT-b & In21k+In1k & 28.7 & \textbf{32.3} & 14.4 \\
         AVT-h \cite{girdhar2021anticipative} & AVT-b & In21k & \textbf{30.2} & 31.7 & \textbf{14.9} \\
         AVT-h \cite{girdhar2021anticipative} & irCSN152 & IG65M & 25.5 & 28.1 & 12.8 \\
         \hline
         \textbf{Ours (Implicit)} & \underline{BNInc} & \underline{In1k} & 28.4 & 31.3 & 14.5 \\
         \textbf{Ours (Explicit: TB)} & \underline{BNInc} & \underline{In1k} & 28.7 & 31.4 & \textbf{14.8} \\
         \textbf{Ours (Explicit: CTP)} & \underline{BNInc} & \underline{In1k} & 28.1 & 31.2 & \textbf{14.8} \\
    \end{tabular}
    \label{tab:ek100}
\end{table}

\subsection{Datasets}
We evaluate our methods on the EPIC-Kitchens-55 (EK55, \cite{atsn}) dataset and its extension EPIC-Kitchens-100 (EK100, \cite{Damen2020RESCALING}). EK55 is a large scale egocentric video dataset with 55 hours of video recordings captured by 32 subjects in 32 kitchens. The data split for the action anticipation tasks is inherited from \cite{DBLP:conf/iccv/FurnariF19}, containing 23492 segments for training and 4979 for validation. All combinations of 125 verbs and 352 nouns in the training split result in 2513 verb-noun action categories. EK100 extends EK55 to 100 hours of video recordings with revised annotations. The data split contains 67217 segments for training and 9668 for validation. EK100 considers 97 verbs and 300 nouns. Unique verb-noun pairs provide 3087 action categories. 

We evaluate the predictive performance of our model for a range of anticipation intervals $\tau_a$, which indicate up to how many seconds before the action start time $t_s$ a model can access frames of the video recording.

\subsection{Implementation Details}
For our experiments with EK55 and EK100 we use the pretrained BN-Inception from \cite{DBLP:conf/iccv/FurnariF19} as a frame-level feature extractor. 
For sampling the dataset, we select a total of 14 observable frames for each segment with a fixed stride of $\alpha_s=0.25$ (4 fps). All input frames are resized to 256x454 and fed to the proposed model. The output from the readout function is then mean-pooled and fed to an unrolling classifier to obtain verb, noun, and action predictions.  The unrolling classifier is a widely adopted design in action anticipation \cite{DBLP:conf/iccv/FurnariF19, osman2021slowfast, tai2021higher}, and we follow the similar classifier design as prior works. The classifier outputs are supervised with ground-truth labels using cross-entropy. We train the model by summing up cross-entropy from eight anticipation intervals, from $\tau_a=2.00$ to $\tau_a=0.25$ with step size $0.25$. 
RandAugment~\cite{cubuk2020randaugment} is used in all experiments. AdaBelief \cite{DBLP:conf/nips/ZhuangTDTDPD20} in combination with the look-ahead optimizer \cite{NEURIPS2019lookahead} is adopted. Weight decay is set to 0.001. Learning rate is initially set to 1e-4 and cosine annealed to 1e-7 on the last 25\% of epochs. We use 4 $\times$ NVIDIA A100 40GB GPUs for training with batch size 32. We train for 50 epochs. All the experiments are conducted on RGB-only modality.

\subsection{Results}
Table~\ref{tab:ek55} shows the comparison of our proposed model and existing methods on EK55. We report accuracy for eight anticipation times $\tau_a$, from 2.00 to 0.25 seconds with step size 0.25. Some methods are constrained by its design, mostly clip based methods, to only being capable of providing the prediction at one pre-defined time step.  Recurrent modeling has no such limitation, and also shows better performance: our proposed method using implicit edge estimation outperforms most of the current state-of-the-art methods. SF-RU ensembles with different sampling rates achieve higher top-5 action accuracy at $\tau_a=1$, but not at $\tau_a=1.5$ and $\tau_a=0.5$ seconds. Using explicit edge learning strategies, Template Bank (TB) and Class-Token Projection (CTP), shows boosted accuracy at $\tau_a=1$, and also brings significant improvement over mean top-5 action recalls. Table~\ref{tab:ek55_3} includes the comparison to the previous winners of EPIC-Kitchens anticipation challenges. Top-1 and top-5 action accuracy at $\tau_a=1$, which were used as evaluation criteria in the challenge, are reported. Under the same backbone usage, our method shows state-of-the-art performance on both top-1 and top-5 accuracy with explicit edge learning. AVT-h equipped with a strong backbone achieves a higher top-1 score but not in the top-5. EK100 evaluation by mean top-5 action accuracy is reported in Table~\ref{tab:ek100}. Our method achieves overall better performance under the same backbone usage, and is slightly weaker but competitive when compared to AVT with a stronger AVT-b backbone.

Overall, our proposed method shows state-of-the-art performance on both EK55 and EK100. Explicit edge learning with the template bank shows slightly better performance than the class-token projection approach on EK100. Class-token projection, on the other hand, obtains 
better accuracy in different anticipation intervals on EK55.

\begin{table}[t]
\caption{Different bank sizes are set on EK55. All the numbers are conducted at 1\textit{s} anticipate interval.}
    \centering
    \scriptsize
    \begin{tabular}{l|c|ccc|ccc}
    Bank &
    \multicolumn{1}{c}{Top-1 (\%)} &
    \multicolumn{3}{c}{Top-5 Acc. (\%)} &
    \multicolumn{3}{c}{Mean Top-5 Rec. (\%)} \\
    Size & Action Acc. & Verb & Noun & Action & Verb & Noun & Action \\
    \hline
    1 & 13.22 & 78.34 & 47.39 & 31.38 & 44.07 & 45.86 & 12.99 \\
    32 & 12.99 & 78.96 & 47.49 & 31.68 & 43.60 & 46.43 & 12.62 \\
    64 & 13.05 & 78.42 & 47.51 & 31.58 & 42.77 & 46.91 & 13.20 \\
    128 & 13.37 & 79.04 & 48.17 & 31.84 & 44.70 & 47.15 & 13.07 \\
    256 & 13.44 & 78.80 & 48.07 & 31.98 & 44.18 & 48.08 & 13.13 \\
    512 & 13.84 & 78.60 & 46.86 & 32.02 & 43.63 & 46.86 & 13.58 \\
    1024 & 13.31 & 78.38 & 48.23 & 32.08 & 42.76 & 47.75 & 13.34 \\
    2048 & 12.89 & 78.98 & 47.18 & 31.23 & 42.35 & 45.76 & 12.63\\
    \end{tabular}
    \label{tab:ablation_bank_size}
\end{table}

\begin{table}[t]
\caption{Design variations of class token projections on EK55. All the numbers are conducted at 1\textit{s} anticipate interval.}
    \centering
    \scriptsize
    \begin{tabular}{l|c|ccc|ccc}
    Class &
    \multicolumn{1}{c}{Top-1 (\%)} &
    \multicolumn{3}{c}{Top-5 Acc. (\%)} &
    \multicolumn{3}{c}{Mean Top-5 Rec. (\%)} \\
    Tokens & Action Acc. & Verb & Noun & Action & Verb & Noun & Action \\
    \hline
    Global & 13.07 & 78.40 & 47.47 & 31.64 & 42.91 & 47.73 & 12.74 \\
    $V, N$ & 13.60 & 78.74 & 47.59 & 31.96 & 44.96 & 47.19 & 13.61 \\
    $V, N, A$ & 13.66 & 78.54 & 48.39 & 31.46 & 43.88 & 47.46 & 13.40 \\
    \end{tabular}
    \label{tab:ablation_cls_proj}
\end{table}


 
 
\subsection{Bank Size}
Table~\ref{tab:ablation_bank_size} shows the experimental results on varying the bank sizes from 1 to 2048. In the case of bank size being equal to 1, a unique template is always selected globally across different timesteps. It can be viewed as placing a strong regularization to the intermediate graph representations, forcing the vertices to be connected in a specific way. When the bank size is larger than 1, the templates are no longer shared globally and become sample conditioned, which brings flexibility of message routing between vertices. According to the experiment results, when the bank size is greater than 64, top-1 / top-5 accuracy and mean recall improve over the configuration with a single global template. The reasonable choices are between 256 to 1024, with a peak performance at 512. No improvement is obtained with more than 1024 templates. Accordingly, we used bank size 512 throughout this study.

\subsection{Design Variants in Class-Token Projection}
We report on the performance of three Class-Token Projection variants. We modified~\eqref{eq:class-token projection} by using: (i) \textit{Global Token}: $P \otimes P$, where $P$ is the projection of the global token used to present verb and noun; (ii) \textit{$V, N$ Tokens}: $V \otimes N$; (iii) \textit{$V, N, A$ Tokens}: $V \otimes N + A \otimes A$, where $A$ is the projection of the action token. Based on the results shown in Table~\ref{tab:ablation_cls_proj}, we conclude that using individual tokens for verbs and nouns consistently achieves better accuracy.
We obtain similar performance by replacing the classifier inputs (mean-pooled over the vertices) with an additional action token. Based on the overall action accuracy, we used $V,N$ tokens for projection in this study.
\subsection{Parameters and Computation Counts}
Our model implements message, update, and readout functions with only a multi-head self-attention (MHSA) followed by a feed-forward block (FFN). The total parameters in our model are about $21C^2$, around 21M when $C$=1024. The template bank introduces additional 6.5M parameters for the selector and the stored templates. The class-token projection, on the other hand, introduces additional 0.23M parameters for  class tokens and projections. The total computation, excluding backbone and classifier, is about 40 GFLOPs per timestep.

\section{Conclusion}
We presented a unified recurrent modeling which generalizes the recurrence mechanism by transferring the sequence into graph representation via message passing framework. We also propose novel edge learning strategies which explicitly approximate the edge connectivity of graph representation. On the large-scale egocentric EPIC-Kitchen dataset we outperform the current state-of-the-art in video action anticipation. The proposed model is generic, utilizes only self-attention, and hence provides a flexible framework that can be further extended. We plan to integrate multi-modality and explore the rich information in the annotations in future work.

\bibliographystyle{IEEEtran}
\bibliography{IEEEabrv,egbib}

\end{document}